\RequirePackage{fix-cm}
\documentclass[a4paper,10pt]{article}
\usepackage{fcds}

\usepackage{graphicx}
\usepackage{comment}
\usepackage{float}

\usepackage{amsmath}
\usepackage{amssymb}
\usepackage{amsthm}
\usepackage{mathptmx}

\usepackage{fcds}
\usepackage{anyfontsize}
\usepackage{t1enc}
\usepackage[labelsep=period]{caption}
\usepackage{enumitem}
\usepackage{titlesec}
\titlelabel{\thetitle.\quad}
\usepackage[top=5.5cm, bottom=5cm, left=4cm, right=4cm]{geometry}
\graphicspath{ {./pics/} }

\date{}

\theoremstyle{definition}
\newtheorem{definition}{Definition}[section]

\graphicspath{{figures/}}

\usepackage{tikz} % checkmark sign
\newcommand{\cc}{\tikz\fill[scale=0.4]
(0,.35)--(.25,0)--(0.7,.7)--(.25,.15)--cycle;}

\usepackage[disable]{todonotes}

\usepackage[pdftex, pdfauthor={Maciej A. Czyzewski, Artur Laskowski, Szymon
Wasik}, pdftitle={Chessboard and Chess Piece Recognition With the Support of
Neural Networks}]{hyperref}

\title{\textbf{Chessboard and Chess Piece Recognition With the Support of Neural Networks}}

\newcommand\blfootnote[1]{%
  \begingroup
  \renewcommand\thefootnote{}\footnote{#1}%
  \addtocounter{footnote}{-1}%
  \endgroup
}

\usepackage[misc,geometry]{ifsym} 
\renewcommand{\thefootnote}{\fnsymbol{footnote}}

\author{Maciej A. Czyzewski\textsuperscript{$\ast$}, {Artur Laskowski\textsuperscript{$\ast$, $\dagger$, \Letter}}, Szymon Wasik \thanks{Institute of Computing Science, Poznan University of Technology, Piotrowo 2, 60-965 Poznan, Poland}\textsuperscript{,}\thanks{European Center for Bioinformatics and Genomics, Poznan University of Technology, Piotrowo 2, 60-965 Poznan, Poland}}
\begin{document}

% FIXME %%%%%%%%%%%%%%%%%%%%%%%%%%%%%%
% \date{Received: date / Accepted: date}

\maketitle

\begin{abstract}
Chessboard and chess piece recognition is a computer vision problem that has not
	yet been efficiently solved. However, its solution is crucial for many
	experienced players who wish to compete against AI bots, but also prefer to
	make decisions based on the analysis of a physical chessboard. It is also
	important for organizers of chess tournaments who wish to digitize play for
	online broadcasting or ordinary players who wish to share their gameplay
	with friends. Typically, such digitization tasks are performed by humans or
	with the aid of specialized chessboards and pieces. However, neither
	solution is easy or convenient. To solve this problem, we propose a novel
	algorithm for digitizing chessboard configurations.

We designed a method that is resistant to lighting conditions and the angle at
	which images are captured, and works correctly with numerous chessboard
	styles.  The proposed algorithm processes pictures iteratively. During each
	iteration, it executes three major sub-processes: detecting straight lines,
	finding lattice points, and positioning the chessboard.  Finally, we
	identify all chess pieces and generate a description of the board utilizing
	standard notation.  For each of these steps, we designed our own algorithm
	that surpasses existing solutions.  We support our algorithms by utilizing
	machine learning techniques whenever possible.

The described method performs extraordinarily well and achieves an accuracy over
	$99.5\%$ for detecting chessboard lattice points (compared to the $74\%$ for
	the best alternative), $95\%$ (compared to $60\%$ for the best alternative)
	for positioning the chessboard in an image, and almost $95\%$ for chess piece
	recognition. \blfootnote{\Letter~artur.laskowski@cs.put.poznan.pl}

\end{abstract}


\section{Introduction} %%%%%%%%%%%%%%%%%%%%%%%%%%%%%%%%%%%%%%%%%%%%%%%%%%%%%%%%%

It is a natural behavior for people to simplify repetitive tasks because we
typically prefer to focus on challenges that require high creativity.
Therefore, machines are widely used for such automatable tasks.  An example of
such a take could be digitizing physical content.  Digitization is a process
that is often very time consuming, does not require specialized knowledge, and
is still frequently performed or substantially supported by humans.  However,
recent developments in computer vision and artificial intelligence algorithms
have improved the performance of software programs for performing such tasks and
reduced the need for human assistance \cite{Wasik_2015}.

In this article, we focus on the problem of detecting chessboards captured in
digital images. A chessboard is the type of board that is utilized to play the
game of chess.  A board consists of 64 squares (eight rows and eight columns)
and the chess pieces that are placed on a board and can hide certain portions of
that board. The aforementioned application is different from the most common
application of the chessboard detection problem, which is typically utilized in
the context of camera calibration processes \cite{de2010automatic}.  Detecting
chessboards for calibrating cameras is a much easier task because one can assume
specific colors of squares and there are no obstacles placed on the board.
Nevertheless, this application is also an important issue for chess games.  It
can be utilized for loading a chess game state into the memory of an autonomous
robot or artificial intelligence (AI) algorithm for playing chess
\cite{Urting2003MarineBlueAL,Matuszek2011GambitAA,Cour2003AutonomousCR}.  Such
a solution is crucial for many experienced players who wish to compete against
AI bots, but prefer to make decisions based on the analysis of a physical
chessboard instead of its digital representation.  A similar problem occurs when
someone wishes to digitize the play in a chess tournament, such as for a live
online broadcast.  Typically, such a digitization tasks are performed by humans
or with the aid of a specialized chessboard and pieces. However, this is not an
easy or convenient solution. To solve this problem, we propose a novel algorithm
for digitizing chessboard states.  In this article, when referring to the
chessboard detection problem, we consider it as two inseparable steps: detecting
the board itself and then detecting and recognizing the types and locations of
chess pieces on the board. Only a method that implements both of these steps can
fully support chess players.

Furthermore, it is worth noting that there are other computer vision problems
that are indirectly connected to the problem of detecting chessboards.  For
example, the recognition of features characterizing chessboards (lines and grid
points) is applied in autonomous vehicles systems \cite{Pomerleau1996}, which
can identify road situations and make appropriate decisions based on the
information they receives via optical sensors \cite{Li2004, Leonard1990}.
Another significant application that could directly benefit from work on
detecting chessboards, particularly the step for recognizing chess pieces, is
face recognition \cite{ZIAUL}.  Face recognition algorithms are widely used in
various security applications, such as authenticating individuals
\cite{Larson2018}.

The problem of recognizing a chessboard from an image is difficult, with a major
issue being the quality of images (e.g., lighting issues and low resolutions for
internet streaming).  The topic of image quality and lighting conditions was
thoroughly studied in the context of face recognition by Braje et al.
\cite{braje1998illumination} and Marciniak et al. \cite{Marciniak2013}.  Most
of the issues discussed in the above papers are also relevant to the chessboard
detection process.  For this reason, most current methods for chessboard
recognition typically perform certain simplifications. This includes utilizing
only a single chessboard style during experiments, capturing images with a
direct overhead view of the chessboard or at a convenient angle specified in
advanced, utilizing specially designed chessboards, such as boards with special
markers that aid in detecting chessboard corners, and other more sophisticated
methods for simplifying the detection task.

In this paper, we propose a novel algorithm for detecting chessboards and chess
pieces that is robust to lighting conditions and image capture angles, and
performs extraordinarily well on a variety of chessboard styles. The exact
colors of board fields are not important, meaning a board can be damaged or
certain obstacles can obscure an image without degrading detection performance.
Later in this paper, we present a comparison between our method and other
approaches that are currently widely used.  Finally, we present a novel enhanced
dataset that can be utilized in any future work on chessboard recognition.

It should be noted that many standard algorithms in the computer vision field
are not currently able to process many real-world scenarios.  They are
successful on standard benchmarks, but do not succeed in many non-trivial cases.
When designing our algorithm, we decided to focus on its applicability in the
real world and adapted all methods utilized to handle any conditions.  This
required us to design our algorithms to be non-parametric, self-adjustable, and
accurate. Furthermore, we wished to limit execution time to less than five
seconds for a single image to make it possible to utilize the algorithm during
chess tournaments to process pictures during live games. For these reasons, we
were forced to create replacements for various modules, such as line and lattice
point detectors, and enhance them with machine learning techniques. As a result,
we created a method that significantly outperforms all existing solutions.  In
the first few weeks after its publication on GitHub, it has already been noted
by many prominent researchers and companies.

\section{Related work} %%%%%%%%%%%%%%%%%%%%%%%%%%%%%%%%%%%%%%%%%%%%%%%%%%%%%%%%%

Many computer vision researchers have attempted to solve the challenging problem
of chessboard detection. As early as 1997, Soh recognized that detecting an
enhanced board is a trivial problem \cite{Soh1997}. Utilizing markers or any
other board-enhancing system improves the ability of algorithms to detect
boards. However, such enhancements significantly decreases algorithm
versatility. However, more generic approaches tend to suffer from environmental
effects (e.g., poor lighting). For this reason, there have been several attempts
to design dedicated hardware for supporting chessboard detection. In 2016, Koray
et al. proposed a real-time chess game tracking system
\cite{koray2016computer}. This system utilizes a webcam positioned over the
chessboard.  Other solutions involve the preparation of a special magnetic
chessboard and set of chess pieces \cite{CoolThings2016}.  However,
such approaches are expensive and difficult to implement.  Computer vision
systems provide an excellent alternative because they are cheaper and relatively
reliable if calibrated properly.

Most computer vision methods work by adapting and combining known and commonly
used transformations and detectors. For example, De la Escalera and Armingol
proposed a method utilizing Harris and Stephens's corner detection method
\cite{harris1988combined} followed by a Hough transform \cite{Duda1972}.  The
latter step was utilized to enforce linearity constraints and discard responses
from the corner detector that did not fall along the chessboard lines because
the authors were aware of defects in Harris and Stephens's detector, which is
likely to find many corners outside the board \cite{de2010automatic}. The ChESS
algorithm was designed to compete with Harris and Stephens's detector in the
field of chessboard recognition \cite{bennett2014chess}.  However, this
geometric detector can process only simple cases.  It encounters difficulty in
recognizing lattice points located near obstacles, meaning each chess piece
decreases the algorithm's capability for detecting lattice points.  The main
advantage of ChESS is the speed of processing a single image. 

One of the most accurate algorithms (up to 90\% of accuracy) was proposed by
Danner and Kafafy \cite{danner2015visual}. However, they simplified the problem
by utilizing only a green-red chessboard, which is inapplicable in nearly all
real-world scenarios.  Many other researchers have proposed corner-based
approaches, such as \cite{arca2005corner} or \cite{zhao2011automated}, which
averaged the quality of results. Relatively fewer researchers have utilized
exclusively line-based methods \cite{Soh1997}, \cite{kanchibailchess}. The
latter method, proposed by Kanchibail, emphasizes one of the weaknesses of these
methods, which is sensitivity to changes in image orientation.  There have also
been attempts to design methods that require additional user interactions, such
as asking a user to manually select the four corners of the chessboard
\cite{dingchessvision}.

Finally, as discussed in the introduction, there are a variety of methods that
utilize chessboard detection algorithms for calibrating cameras. One such method
and a well-written review of the most commonly used approaches was published by
Zhang \cite{zhang2000flexible} and later updated by Shortis in the context of
underwater cameras \cite{Shortis_2015}.  Additionally, Tam et al. introduced
the classification of such methods into line-based and corner-based approaches
\cite{tam2008automatic}.  However, the capabilities of such methods are too
limited to be applied to recognizing chessboards with chess pieces on them.

\section{Materials and methods} %%%%%%%%%%%%%%%%%%%%%%%%%%%%%%%%%%%%%%%%%%%%%%%%%%%%%

\begin{figure*}[t]
    \centering
	\includegraphics[width=\textwidth]{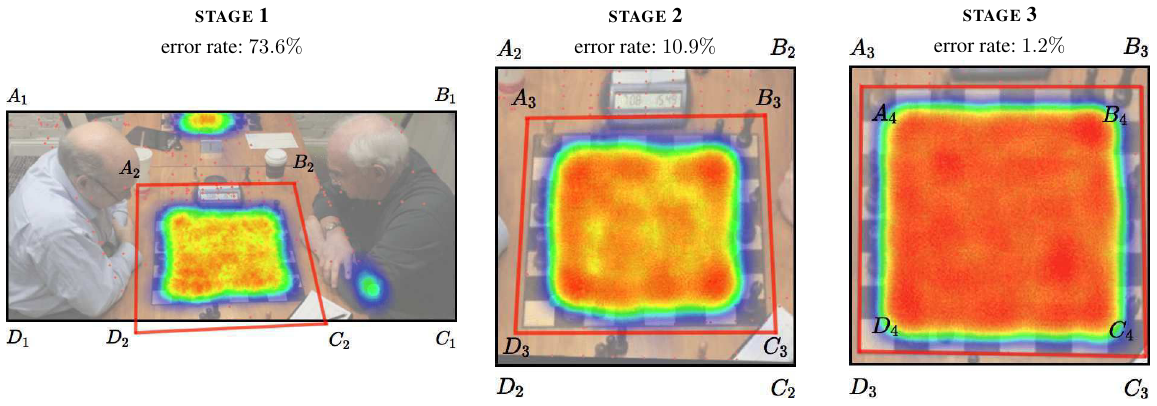}
	\caption{Demonstration of how the lookup process works. In this case, we
	have three stages. The final stage precisely locates the corners of the
	chessboard (i.e., $A_4 B_4 C_4 D_4$).  It is worth noting that during the
	first stage, the algorithm also finds a neighboring chessboard. A definition
	of the error rate can be found in Section \ref{ssec:stage}.}
	\label{f:1}
\end{figure*}

The main objective of our research was to design a method for locating and
cropping a chessboard in an image, which could then be utilized to create a
digital record of chess pieces positions utilizing Forsyth-Edwards notation
(FEN) \cite{edwards1994portable}, which is the most commonly used notation for
representing states of chess games.

Precise chessboard positioning within images is a computer vision problem of
extraordinary difficulty. It requires accurate locating of the four edges of a
chessboard. It is a very computationally complex process to precisely locate
these edges in a source image in a single step. For this reason, our solution is
based on iterative heat map generation.  The generated heat map visualizes the
probability that the chessboard is located in a specific sub-area of an image.
After generating the heat map matrix, we assume that the chessboard is located
in the tetragonal sub-area containing the highest probability values. We then
crop this sub-area and correct its perspective to create a square image and
iteratively repeat this procedure until the solutions converge. Hereafter, a
single iteration as described above will be referred by the term \textit{stage}.
Several example outputs of the consecutive stages are presented in Figure
\ref{f:1}.

The use of a heat map image improves the quality of results and reduces
processing time, which is crucial for real-time applications. The iterative
process that we implemented mimics the natural mental process that the human
brain utilizes.  To locate corners of an object, a human first focuses on the
entire object by rotating their head to the correct orientation, and then find
exact points by performing eye movements.  Such an approach can be easily
adapted for processing images containing many chessboards. It simply requires
the inclusion of a clustering technique that can locate all chessboards during
the first stage of the algorithm.  However, this is a trivial generalization, so
this paper focuses on the detection of a single chessboard.

In the following subsections, we first present the algorithm that is executed
during each stage of image processing (Section \ref{ssec:stage}). This algorithm
consists of three major sub-procedures that are executed consecutively to locate
a chessboard in an image.  These sub-procedures are detecting straight lines
(Section \ref{ssec:SLID}), finding lattice points (Section \ref{ssec:LAPS}), and
constructing heat maps (Section \ref{ssec:CPS}). When the final stage of the
algorithm is completed, it outputs a cropped image of the chessboard with
corrected perspective.  This picture is utilized for locating and classifying
chess pieces and generating a description of the board utilizing FEN notation.
This process is detailed in the final subsection of the methods description
(Section \ref{ssec:FEN}).

\subsection{Single processing stage}
\label{ssec:stage}

The objective of each stage of the proposed algorithm is to find a better
approximation of the chessboard position in an image. It has been proved that
the problem of localizing objects in images can be successfully solved by
utilizing a deep convolutional neural network that has been trained to perform
image classification tasks. For example, an advanced and very promising method
was presented in \cite{gao2017dual} for polarimetric synthetic aperture radar
image classification. However, this method is only capable of classifying
various objects appearing in arbitrary positions in an image, but not for exact
object positioning.  This technique could be extended, as described in
\cite{bency2016weakly}. However, it requires 16 iterations to locate an object
inside a rectangular frame without accounting for perspective.  Therefore, to
drastically improve the effectiveness of our approach, we decided to design a
hybrid method. We approached the problem by utilizing altered versions of a
classic computer vision methods based on algorithms boosted by neural networks
to solve various sub-problems. Our method finds the characteristic structures in
an image, such as lines and lattice points, and then assesses their locations
and shapes based on a scoring function called polyscore (cf. Section
\ref{sssec:polyscore}).  The values of the polyscore function define the
temperature of each point in the heat map.  Based on these polyscore values, we
can identify components representing a single chessboard as follows:
\begin{definition}[Heat map component]
\label{def:heat-map-comp}
A heat map component is a connected sub-area of the heat map that is (1) as
large as possible and (2) contains only points with a temperature (i.e.,
polyscore value) greater than some threshold $t_h$. 
\end{definition}
At the end of each stage, the algorithm chooses the tetragonal frame from the
heat map containing a single heat map component (in practice, the frame with
maximal polyscore value is selected, cf. Section \ref{sssec:polyscore}), crops
this frame with an additional offset $P$ based on the error rate value $E$, and
warps the perspective. We define
\begin{gather}
	E = 100\% - A/A_0 \label{eq:1a} \\
	P = \max_{} \{\sqrt{EA}, \frac{\sqrt{A}}{6}\} \label{eq:1b},
\end{gather}
where $E$ in Equation \ref{eq:1a} is an error rate and $P$ in Equation
\ref{eq:1b} is an offset value. The variable $A$ determines the area of a given
heat map component and $A_0$ is the area of the entire image. Additionally, we
must consider the compulsory offset in Equation \ref{eq:1b} because our
algorithm finds only the inner lattice points of the chessboard (i.e., a grid
separating $6\times 6$ squares, omitting lattice points located on the edges of
the chessboard).

The number of stages required to process an image depends on the level of
complexity in that image. Formally, stages are repeated iteratively until the
error rate value decreases to below a small threshold value $t_e$. In practice,
one or two stages is typically sufficient to correctly generate a final heat
map.  We divided the problem of generating a heat map into sub-problems that are
solved by separate modules specialized at solving specific sub-tasks.
Specifically, we utilize three principal modules:

\begin{itemize}
\item The straight line detector (SLID) finds straight lines in an image and
	filters out lines that are expected to have no usability for detecting a
		chessboard position. To simplify the recognition of lattice points by
		the next module, we also designed a novel heuristic algorithm that
		merges many collinear segments into a single line.
\item The lattice points search (LAPS) module selects points that have a high
	probability of being a lattice point (i.e., a point where the corners of
		chessboard squares intersect). This module utilizes a neural network to
		facilitate a geometric detector for recognizing difficult cases of
		lattice points.
\item The chessboard position search (CPS) module computes a heat map
	representing the probability that a chessboard is located in a specific
		sub-area of an image (see Figure \ref{f:1}). The heat map is computed
		primarily based on the results of the SLID and LAPS modules.
\end{itemize}

%\begin{figure}\centering
%	\includegraphics[width=\columnwidth]{2_diagram}
%	\caption{A single stage is defined by the above workflow. On the output we
%	receive a heat map from which we extract the best area for the next stage.}
%	\label{f:2}
%end{figure}

\subsection{Detecting straight lines}
\label{ssec:SLID}

\begin{figure*}[t]
    \centering
	\includegraphics[width=\textwidth]{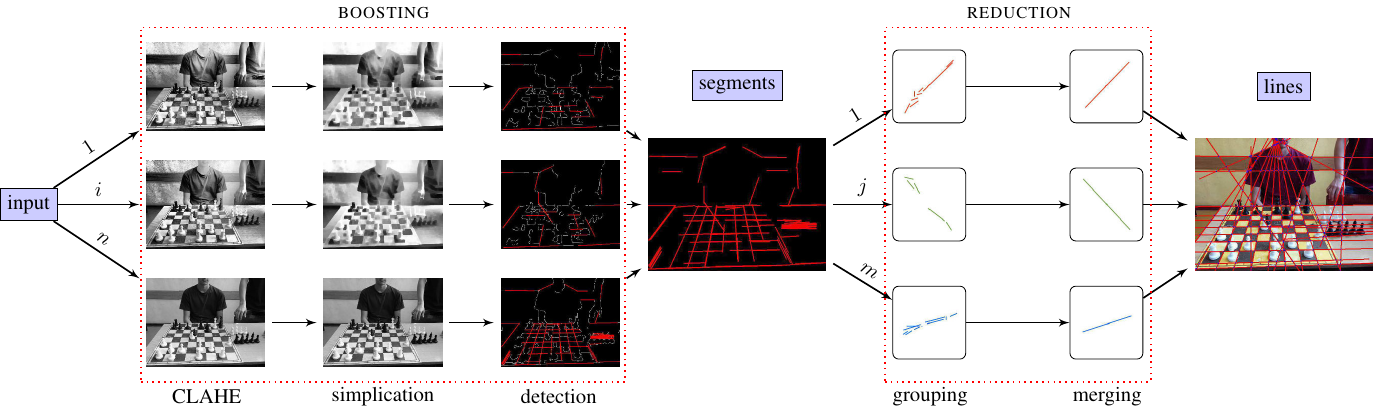}
	\caption{The process of straight line detection. We preprocess images
	utilizing four sets of parameters (i.e., $n=4$), each which is utilized to
	find a portion of the line segments that are then divided into $m$ groups
	and merged into $m$ straight lines ($m$ denotes the number of groups of
	collinear segments and is set automatically by the algorithm).}
	\label{f:4}
\end{figure*}

Line segment detection is a classical problem in image processing and computer
vision.  SLID is an extension of the standard line detector. Its additional
objective is to merge all small segments that are nearly collinear into long
straight lines.  Therefore, there we utilize a standard line detector to
detect segments that can be extracted from an image and merged later.  There
are several line detectors that can be applied to solve this problem.  The
Hough transform is a traditional line detector based on an edge map
\cite{Duda1972}, but it typically extracts infinitely long lines instead of
line segments and typically outputs many false detections, such as those
described by Fernandes and Oliveira \cite{fernandes2008real}.  More
satisfactory results can be achieved by the \textit{Canny Lines} detector
\cite{lu2015cannylines}, which is the detector that we decided to utilize
in our method.

The problem of segment merging has been described by Tavares and Padilha
\cite{tavares1995new}.  They presented a solution based on the analysis of
the centroids defined by two segments. Unfortunately, this algorithm can
generate results that are unintuitive for humans, who typically assume that
a merged line should be as collinear as possible with the longest segment.
This factor is also omitted by many other available algorithms. One
interesting alternative is the method designed by Hamid and Khan
\cite{hamid2016lsm}, which is more perceptually accurate.  However, it is a
slow method with a computational complexity of $O(n^3)$.  Additionally, it
only links segments that overlap or are very close to each other, meaning it
does account for broken lines (i.e., two segments that are collinear, but
are far apart from each other).

Our proposed SLID algorithm consists of three main steps (for a visual
representation, see Figure \ref{f:4}):

\begin{enumerate}
\item Boosting: find all possible segments utilizing multiple analysis of the
	same image via the gradiental threshold method proposed by Sen and Pal
		\cite{sen2010gradient} and various contrast limited adaptive histogram
		equalization (CLAHE) masks \cite{reza2004realization,Stark_2000}.
\item Grouping: separate segments into groups of nearly collinear segments
	utilizing the \textit{linking function} (cf. Section \ref{sssec:link-fun}).
\item Merging: analyze and merge the segments in each group utilizing the
	M-estimator, resulting in one normalized straight line.
\end{enumerate}

\subsubsection{Boosting input}

One effective method for boosting the detection of line segments is to
adaptively adjust the low and high thresholds of the Canny operator based on the
gradient magnitude of the input image. This method can ensure the completeness
of an image's structural information and it was utilized in the CannyPF
algorithm \cite{lu2015cannylines}. Our method is similar, but we also utilize
the CLAHE algorithm \cite{reza2004realization} and a simplification phase based
on erosion and dilation operations to reduce noise and remove insignificant
details.  Our input images pass through a pipeline with different parameters for
CLAHE, erosion, and dilation functions (fig. \ref{f:4}). We determined that to
extract all structural information, one must only analyze four different cases,
namely parameter sets selected manually to correct low light, overexposure,
underexposure, and blur.

\subsubsection{Linking function}
\label{sssec:link-fun}

\begin{figure}[H]
    \centering
	\includegraphics[width=\columnwidth]{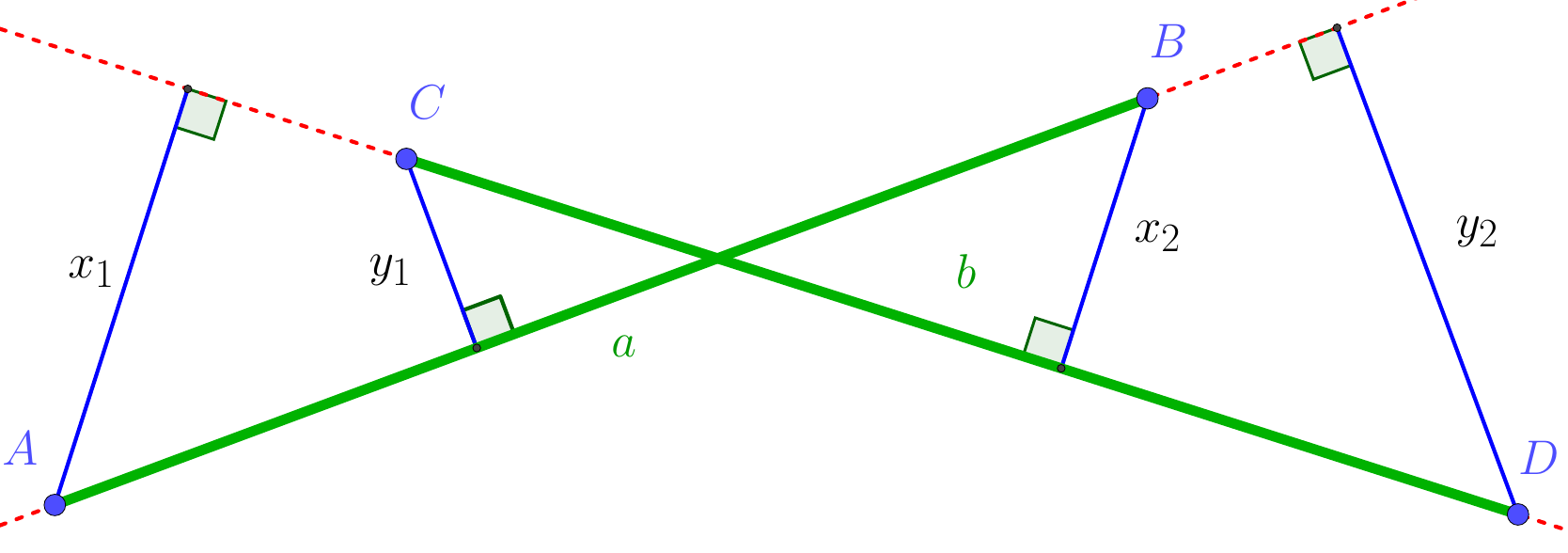}
	\caption{Statistics calculated for each pair of line segments detected in
	the image.}
	\label{f:5}
\end{figure}

The linking function analyzes two segments $\overline{AB}$ and $\overline{CD}$
and attempts to determine if they are located and oriented in such manner that
they can be treated as a single straight line segment. To this end, it utilizes
two heights $x_1$ and $x_2$, which are the lengths of perpendiculars dropped
from points $A$ and $B$ to the line $\overleftrightarrow{CD}$ (cf., Figure
\ref{f:5}) and two heights $y_1$ and $y_2$, which are calculated similarly for
the line $\overleftrightarrow{AB}$.  Additionally, we denote $a = |AB|$, $b =
|CD|$, and
\begin{gather}
	\Delta=(a+b)\cdot t_{\Delta}\label{eq:1}\\
	\gamma=\frac{1}{4}(x_1+x_2+y_1+y_2)\label{eq:2}.
\end{gather}
Here, $\Delta$ is the allowed matching error, $\gamma$ defines the average
deviation from the straight line, and $t_{\Delta}$ represents the acceptable
differences in the positions and orientations of lines according to Equation
\ref{eq:p2}.

Based on the variables defined above, we can define a condition that determines
if two lines should be merged as follows:
\begin{gather}
	\left(\frac{a}{\gamma}>\Delta\right)
	\wedge\left(\frac{b}{\gamma}>\Delta\right)\label{eq:3}
\end{gather}
The parameter $t_{\Delta}$ is defined based on two other parameters, namely $p$,
which defines the degree to which segments should be similar, and $\omega$,
which is a scale constant defined with based on the area of the entire image
$A$. Specifically,
\begin{gather}
	t_{\Delta} = p\cdot\omega \label{eq:p2}\\
	\omega = \frac{\pi}{2}\frac{\sqrt{2\sqrt{A}}}{\sqrt{2A}} =
	\frac{\pi}{2}\frac{1}{\sqrt[4]{A}} \label{eq:p1}.
\end{gather}
We assume that it is relatively difficult to determine if two segments whose
lengths are smaller than the square of the image size are collinear. For this
reason. the image area $A$ in the numerator of the Equation \ref{eq:p1} is under
a double square root.  For example, in our study, pictures were always scaled to
have the same size of ($500 \times 500$ px = $250\,000$ px) and we wished to
connect segments that were similar with a value of $p=90\%$.  Therefore, the
calculation of $t_{\Delta}$ was performed as follows:
\begin{gather}
	\omega = \frac{\pi}{2}\frac{1}{\sqrt[4]{500\cdot500}} \approx 7\%
	\label{eq:p3}\\
	t_{\Delta} = 90\%\cdot7\% = 6.3\% \label{eq:p4}
\end{gather}
$t_{\Delta}$ is an important parameter because if segments are short compared to
the height or width of the entire image, even if they do not appear collinear,
they should be treated as one line. However, the longer these lines are, the
stricter the requirement for their similarity should be. This models the
real-life visual correctness (perceptual accuracy) of line similarity, which
also depends on the scale that the observer sees.  

Grouping can be implemented utilizing a disjoint set data structure to achieve
the best computational complexity possible \cite{Galler_1964, Tarjan_1975}.  We
simply iterate over pairs of segments and if the linking function determines
that a pair is collinear, we perform a union of the sets that contain each
segment.

\subsubsection{Merging}

\begin{figure}[H]
    \centering
	\includegraphics[width=\columnwidth]{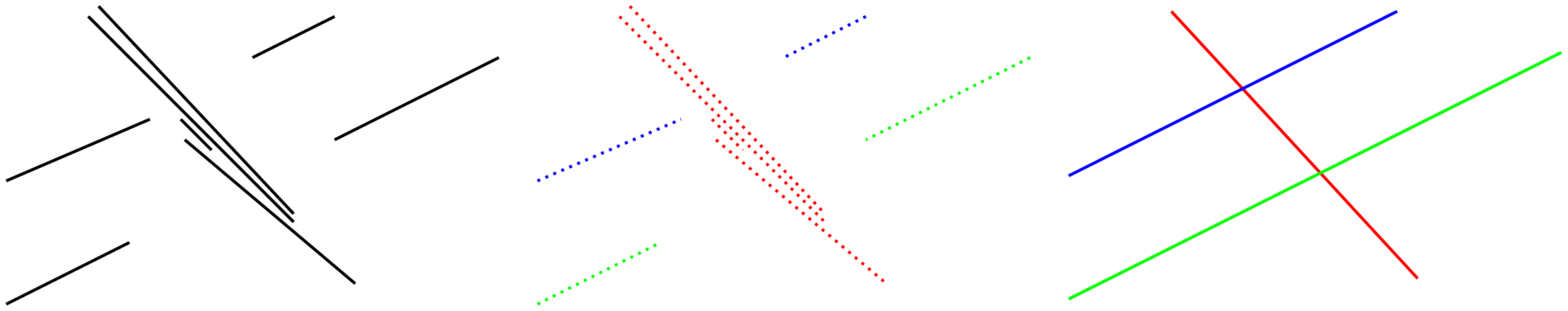}
	\caption{Results of grouping and merging operations. Each color represents a
	different group of collinear segments. These groups are converted to points
	and then utilized to fit straight lines.}
	\label{f:5b}
\end{figure}

The final stage of line detection is merging, which is performed for each group
of collinear segments (see Figure \ref{f:5b}).  This process is divided into two
steps: (1) converting segments to points, where the number of points depends on
the length of a given segment, and (2) fitting a straight line to all points.
The fitting algorithm that we utilize is based on the M-estimator of regression
for approximately linear models \cite{wiens1996asymptotics}, that iteratively
fits lines by utilizing the weighted least-squares algorithm to minimize the sum
of squares of the residuals.  However, it should be noted that the type of
estimator utilized does not have a significant impact on the results.

\subsection{Detecting lattice points}
\label{ssec:LAPS}

As described at the beginning of this paper, detection of chessboard lattice
points is a common problem in computer vision. It has many applications (such as
calibrating video systems) and current detectors are fast, and acceptably
accurate and robust for most practical applications.  However, our algorithm
demonstrates that such detectors can be improved even further. Improvements are
particularly important for correctly processing poor-quality images, such as
images with low-contrast, noise, or those many reflections. LAPS is an
experimental, self-learning chessboard vertex detector based on an embedded
neural network and is the direct successor of the \cite{bennett2014chess} ChESS
detector.  Initially, our algorithm assumes that each intersection of any pair
of lines detected by the SLID module can be a chessboard lattice point.  It
processes each of these points utilizing both geometric and neural detectors,
and returns a list of points that it detects to be lattice points.

The LAPS algorithm takes a $21\times 21$ matrix whose elements represent pixels
as an input.  To verify if an $(x, y)$ point in an image is a chessboard lattice
point, it utilizes a sub-image with coordinates ranging from $(x-10, y-10)$ to
$(x+10, y+10)$.  Then, the algorithm preprocesses this matrix utilizing the
following steps: (1) conversion to grayscale, (2) application of Otsu method
altered by Jassim and Altaani \cite{jassim2013hybridization}, (3) application
of Canny detector, and (4) binarization.  These four operations ensure the
completeness of an image's structural information.

The preprocessed matrix is handled by two modules: (1) a simple geometric
detector that recognizes only perfect cases and (2) a neural network for
recognizing deformed and distorted patterns. First, for the geometric detector,
if the result is positive, we assume that it represents a chessboard lattice
point. Otherwise, we utilize the neural network detector because its result
definitively determines if the matrix represents a chessboard lattice point.

%\begin{figure}\centering
%	\includegraphics[width=\columnwidth]{8_laps}
%	\caption{The diagram illustrates the relations between the geometric and neural detector.}
%	\label{f:6}
%\end{figure}

\subsubsection{Geometric detector}

\begin{figure}[H]\centering
	\includegraphics[width=\columnwidth]{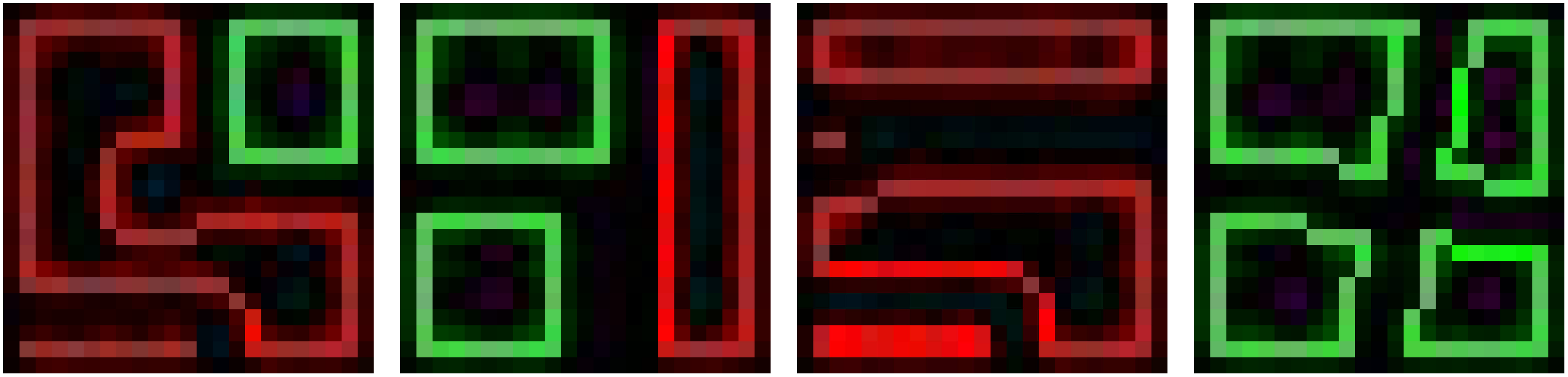}
	\caption{Example results of the geometric detector. Green color represents
	rhomboids that were identified by the detector. Only the rightmost matrix
	will be classified as a lattice point because it is the only one that
	contains four rhomboids.}
	\label{f:7}
\end{figure}

The geometric detector is very simple, meaning it can only recognize trivial
cases (see Figure \ref{f:7}). This detector utilizes the following algorithm:
(1) add a 1-pixel-width frame of the background color (black) around the input
matrix, (2) perform morphological erosion, (3) find all contours and (4) check
if the contour resembles a rhomboid. If there are four rhomboids detected in an
image, it means that the matrix contains a chessboard lattice point because the
rhomboids correspond to four quadrants that are separated by crossing lines.

\subsubsection{Neural detector}

As already described earlier in this section, a $21\times 21$ matrix is given as
the input for the neural network detector.  We based our neural detector on a
convolutional neural network consisting of two layers: a convolutional 2D layer
with 12 filters and a flattened layer with a $0.5$ dropout function.  The neural
network has two outputs with values between zero and one.  Such a design is
commonly used for binomial classification.  In our application, there are two
classes denoting if a matrix represents a chessboard lattice point (i.e. classes
of positive and negative samples).

It should be noted that we analyzed images that were downscaled to a size of
$500\times 500$ pixels.  Therefore, the input matrices for our network represent
$0.176\%$ of the input images and the detector can learn how other objects
overlap with the chess lattice points. For example, our detector can recognize
cases where there is a chess lattice point with half of the input matrix hidden
behind a pawn's head (appearing as a circle with two tangents). 

Our convolutional neural network was trained on a few thousand images of
difficult chess lattice points and other patterns resembling lattice points
(4,732 positive and 4,933 negative samples). To generate such a large training
dataset, we utilized an automated procedure that modifies perfect, artificial
images of 2D standard $8\times 8$ chessboards by warping their perspective in a
random direction. In this manner, we prepared 10\% of the initial dataset.
Additionally, during a training session on real photographs at the end of the
process of detecting chessboard positions, we removed all chessboard lattice
points (even those that were not detected properly). We then added them to the
dataset as positive or negative samples, but only if the current pre-trained
network was certain that a sample was positive or negative with a confidence
value over 75\%. All other samples were ignored.  In this manner, we ignored all
positive cases in which an object was blocking a clear view (player hands,
pawns, or other item) and all negative cases that had some chance to be a
lattice point. As a result, we generated a dataset with very difficult, damaged,
and deformed samples, which increased the capabilities of our neural detector.

Our dataset is accessible from the RepOD repository under the name LATCHESS21
\cite{Czyzewski_2018}, where we published it for open access
\cite{Szostak_2016,Mietchen_2018}.  The neural network and pre-trained model
are available in the supplementary materials.

\subsection{Searching for chessboard positions}
\label{ssec:CPS}
\label{sssec:polyscore}

The CPS algorithm searches for chessboard positions by analyzing a four-sided
frame that can enclose a chessboard. For this purpose, it selects four lines
detected by the SLID algorithm that form a quadrilateral, which is later scored.
The main problem is how to avoid $n \choose 4$ operations. It turns out that it
is trivial to optimize the algorithm by limiting it to a few cases. In
particular, this algorithm utilizes the following operations:

\begin{enumerate}
\item Find clusters of lattice points generated by the LAPS algorithm and choose
	the group with the largest number of points. We denote this group as $G$.
		It should represent the primary chessboard in a picture.
\item Calculate $\alpha = \frac{\sqrt{A_G}}{7}$ and the group centroid, where
	$A_G$ determines the surface area of a group $G$. Based on this formula,
		$\alpha$ will approximate the width of a chessboard square.
\item For each chessboard lattice point in the group $G$, find the lines
	generated by the SLID algorithm that satisfy the following conditions:
	\begin{itemize}
		\item their distance from the closest chessboard lattice point is at
			most $\alpha$;
		\item their distance from the group $G$ centroid is at least $2.5 \cdot
			\alpha$, which will remove lines that are in the central area of the
			group;
		\item they have a high chance of being near a frame edge (to verify
			this, we utilize a polyscore function (see Equation \ref{eq:6}) and
			check if its value is not equal or almost equal to 0).
	\end{itemize}
\item Divide the lines identified in the previous step into two groups, namely
	horizontal and vertical lines, while accounting for perspective.
\item Take pair of lines from each group (two horizontal and two vertical lines)
	to form a frame and calculate the polyscore of each frame. Finally, choose
		the frame that maximizes the polyscore.
\end{enumerate}

\begin{figure}[H]
    \centering
	\includegraphics[width=\columnwidth]{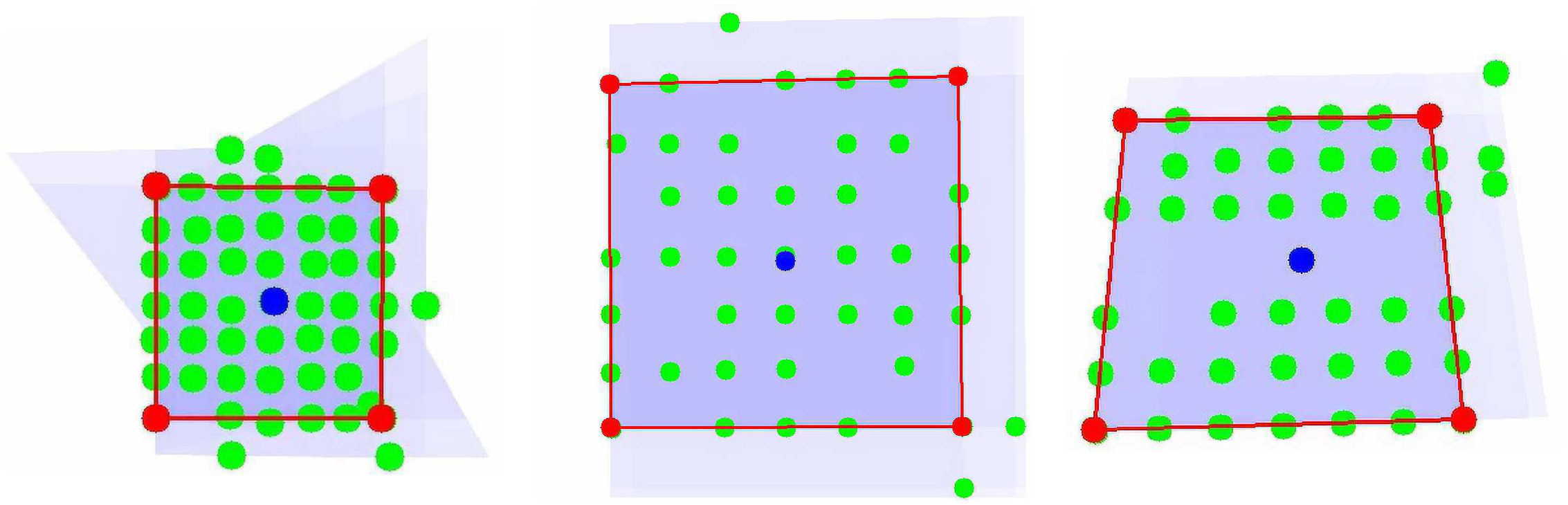}
	\caption{Polyscore values calculated for various frames. The frame with the
	highest score is marked with the red border. Higher scores are highlighted
	with darker blue backgrounds.  Based on the average values of the polyscore
	functions for the frames covering each pixel, we can draw the heat maps
	presented in Figure \ref{f:1}.}
	\label{f:8}
\end{figure}

In order to analyze the probability that a frame $F$ defined by four lines
encloses a chessboard, we utilize the scoring function $P(F)$, which is called a
\textit{polyscore} (see Equation \ref{eq:6} and Figure \ref{f:8}).  This is an
improved density function for chessboard lattice points in a given map segment:
\begin{gather}
W_i(x) = \frac{1}{1+\left(\frac{x}{A_F}\right)^{\frac{1}{i}}}\label{eq:5}\\
P(F) = \frac{L^4}{A_F^2} \cdot W_{3}(k) \cdot W_{5}(l)\label{eq:6}
\end{gather}
Here, $W_i(x)$ is a weight function that applies a weight $i$ with respect to
the area of the frame $F$, which is denoted $A_F$, where $L$ is the number of
points inside the frame, $k$ is the average distance of points inside the frame
from the nearest side of this frame, and $l$ is the distance between the group
$G$ centroid and frame $F$ centroid. The function $P(F)$ has a maximum value for
a frame $F$ representing a perfectly cropped chessboard.

\subsection{Forsyth-Edwards Notation generation}
\label{ssec:FEN}

\begin{figure}%[H]
	\centering
	\includegraphics[width=\columnwidth]{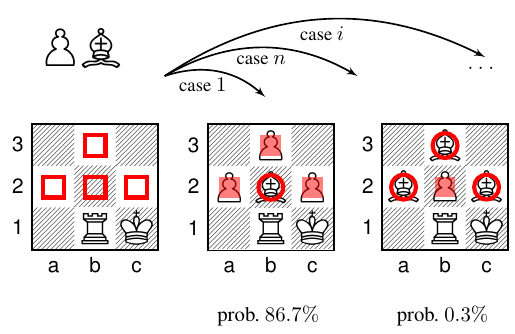}
	\caption{An example situation illustrating cases that we can easily
	discard as improbable by utilizing a chess engine, such as Stockfish.}
	\label{f:3}
\end{figure}

After finding the position of a chessboard, the algorithm proceeds to the phase
of recognizing chess piece located on the chessboard. The most popular approach
for solving this problem is based on utilizing color segmentation to detect
pieces and shape descriptors to identify them. However, this method provides
unsatisfactory results because chess figures from a bird's-eye view are nearly
indistinguishable. In our study, we utilized a similar method described by Ding
\cite{dingchessvision}, which achieves approximately $90\%$ accuracy for
individual chess pieces.  We attempted to utilize both the original support
vector machine designed by Ding and an alternative convolutional neural network
that we designed. The accuracies of both methods were similar. However, we
discovered that we could increase accuracy significantly by implementing the
following major improvements:

\begin{enumerate}
\item Utilizing a chess engine: We utilized the open-source Stockfish engine,
	which allows us to calculate the most probable piece configurations. By
		calculating the probabilities of all possible configurations, we could
		choose the most probable candidate (see Figure \ref{f:3}).
		Additionally, we strengthened this method by utilizing large-scale chess
		game statistics in the manner proposed by Acher and Esnault
		\cite{acher2016large}.
\item Clustering into groups: After clustering similar figures into groups, we
	can reject certain trivial cases based on the cardinality of clusters.  For
		example, having two groups of similar figures with cardinalities of
		$\{5,1\}$ and candidates of $\{\text{bishop},\text{pawn}\}$, it can be
		deduced that there are five pawns and one bishop.
\item Considering physical properties: We utilized the height and area of chess
	figures as additional parameters for classification.  A similar mechanism
		was described by Wu et al. \cite{wu2018alive}.
\end{enumerate}

We determined that the effectiveness of the piece detector became less important
after implementing the improvements described above. In fact, we only require a
hypothesis regarding which piece could be located in a given square because our
module deduces the most probable situation.

\section{Results} %%%%%%%%%%%%%%%%%%%%%%%%%%%%%%%%%%%%%%%%%%%%%%%%%%%%%%%%%%%%%%

For testing our algorithm, we wished to prepare a challenging benchmark dataset
consisting of images of chessboards captured under various, difficult
conditions. To collect images containing chessboards and chess pieces that were
difficult to recognize, we defined the following conditions for our benchmark
images:
\begin{itemize}
\item there are objects in the picture other than the chessboard and chess
	pieces;
\item straight lines in the image are not generated solely by the chessboard,
	but also by other objects;
\item the chessboard is not perfect (e.g., computer graphics), meaning it does
	not fill an entire image and the corners and chess pieces are not emphasized
		in any way;
\item there are shadows, reflections, distortions, and noise in image.
\end{itemize}

\begin{table*}[t]
\centering
\begin{tabular}{l|*{10}{c}|r}
\multicolumn{1}{l}{Method} &
\multicolumn{10}{c}{Chessboard ID} &
\multicolumn{1}{r}{Accuracy} \\
\hline
                        & 1 & 2 & 3 & 4 & 5 & 6 & 7 & 8 & 9 &10 &\\
\hline
Our                     &\cc&\cc&\cc&\cc&0.5&\cc&\cc&\cc&\cc&\cc&95\%\\
\cite{de2010automatic} &   &\cc&   &   &   &\cc&\cc&\cc&\cc&\cc&60\%\\
\cite{danner2015visual}&   &   &   &   &   &\cc&\cc&\cc&   &\cc&40\%\\
\cite{dingchessvision} &-  &-  &-  &-  &-  &-  &-  &-  &-  &-  & N/A\\
\end{tabular}
\caption{Comparison of chessboard detection performances. A 0.5 score indicates
	that the chessboard was cropped properly, but the corners were not perfectly
	matched for various reasons.  The main objective of Ding's research
	\cite{dingchessvision} was to introduce a novel approach for chess piece
	recognition and the authors assumed that a user would manually select the
	four corners of the chessboard.}
\label{t:1}
\end{table*}

\begin{table*}[t]
\centering
\begin{tabular}{l|*{10}{c}|r}
\multicolumn{1}{l}{Method} &
\multicolumn{10}{c}{Chessboard ID (number of pieces)} &
\multicolumn{1}{r}{Error} \\
\hline
& 1 (29) & 2 (24) & 3 (21) & 4 (21) &
  5 (19) & 6 (23) & 7 (15) & 8 (14) &
  9 (31) & 10 (26) & \\
\hline
	Our & \textbf{0} & \textbf{0} & \textbf{2} & \textbf{1} &
		  \textbf{5} & \textbf{2} & \textbf{0} & \textbf{0} &
		  \textbf{2} & \textbf{0} & 1.2p\\
	\cite{dingchessvision}
		& \textbf{0} & 5          & 3          & 4          &
		  23         & \textbf{2} & 2          & 1          &
		  \textbf{2} & 3          & 4.5p\\
	\cite{danner2015visual}
		&            &            &            &            &
		             & 6          & 4          & 3          &
		             & 5          & 4.5p\\
\hline\hline
	Accuracy (best method)
		& 100\%      & 100\%      & 90.47\%    & 95.23\%    &
		  73.68\%    & 91.30\%    & 100\%      & 100\%      &
		  93.54\%    & 100\%      & 94.42\%
\end{tabular}
\caption{Correctness comparisons for chess piece detection. Empty entries in the
	table indicate that an algorithm did not locate a chessboard correctly,
	meaning it was unable to analyze it further. The individual entries indicate
	the numbers of negatively recognized individual squares. The error column
	contains the average numbers of incorrectly detected pieces. The header line
	contains the number of available pieces. For each chessboard we bolded the
	best result.}
\label{t:2}
\end{table*}

Similar to the methodology utilized by Ding \cite{dingchessvision}, we prepared
a benchmark dataset of 30 images of chessboards with a variable number of pieces
placed in randomized configurations. The sources of the images were very
diverse; some were taken by us, some originated from broadcasts of chess
tournaments (we received the consent of the organizers to utilize their images),
and we even included one image scanned from a painting from the thirteenth
century.  To compare our algorithm to other methods, we selected the 10
most-challenging photos from our benchmark dataset. The other twenty photos were
utilized for debugging and training classifiers.

Prior to testing the quality of chess piece recognition, we analyzed how our
chessboard detection algorithm performs compared to other methods. For
comparison methods, we selected the best-performing current methods. The results
are listed in Table \ref{t:1}. One can see that cropping a chessboard from an
image is not a trivial problem. Only our algorithm was able to locate
chessboards in a high percentage of cases.  The other methods tested located no
more than $60\%$ of chessboards correctly.  This issue is often dissembled by
the authors of papers that are focused on chess piece identification. They
simply select test photos for which their algorithm locates a chessboard
correctly. It should be noted that we have not tuned our algorithm for these
test cases because we utilized the other 20 photos from the benchmark dataset to
train our algorithm.

To derive additional insights into why our algorithm is so successful at
locating chessboards, we also compared our chessboard lattice point detector to
the best current alternative, namely the ChESS detector
\cite{bennett2014chess}. As shown in Table \ref{t:3}, our algorithm has very
high accuracy (over $99.5\%$) compared to the $74.3\%$ accuracy of the ChESS
detector. Additionally, it is only slightly slower than the ChESS detector.

\begin{table}[H]
\centering
\begin{tabular}{ l | c | r }
Detector & Accuracy & Time \\
\hline
LAPS     & 99.57$\pm$0.0147\% & 4.57s \\
ChESS    & 74.32\%            & 3.38s
\end{tabular}
\caption{Effectiveness comparisons of chessboard lattice point detectors. The
	final column lists the total times required to process all lattice points
	for all tested chessboards.}
\label{t:3}
\end{table}

Next, we tested the correctness and accuracy of our chess piece recognition
algorithm.  During our tests, we compared our algorithm to the same set of
methods that we utilized to analyze chessboard detection performance, excluding
the approach that was dedicated exclusively to chessboard detection and did not
implement chess piece recognition (i.e., \cite{de2010automatic}). The results
presented in Table \ref{t:2} demonstrate that our method is characterized by
high accuracy and stability for challenging input cases. The method by Danner
and Kafafy \cite{danner2015visual} was unsuccessful when analyzing images of
poor quality and those containing a large number of extraneous objects. The same
problem was identified in the method by Ding \cite{dingchessvision}, which was
misled by images containing too many extraneous lines occurring in regular
patterns. 

\begin{figure}%[H]
	\centering
	\includegraphics[width=\columnwidth]{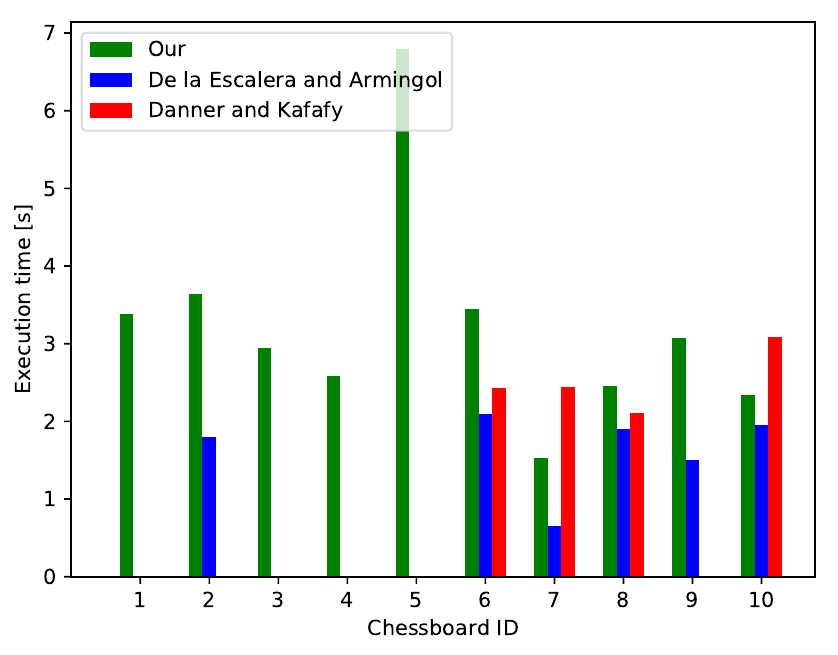}
	\caption{Comparison of the speeds of popular chessboard detection algorithms
	on our benchmark dataset.  When an algorithm could not find the chessboard,
	the processing time was typically much shorter because certain parts of the
	algorithm were omitted. For this reason, we only included processing times
	for cases where a chessboard was successfully located in the source image.}
	\label{f:11}
\end{figure}

\begin{figure}%[H]
	\centering
	\includegraphics[width=\columnwidth]{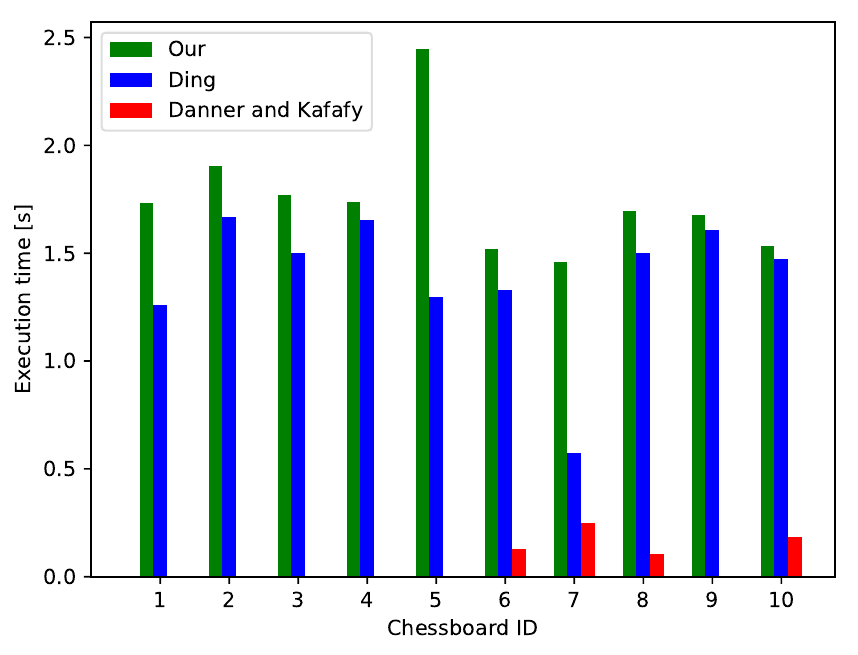}
	\caption{Comparison of the speeds of popular chess piece detection
	algorithms on our benchmark dataset.  When an algorithm could not find the
	chessboard, it was not possible to utilize it for identifying chess pieces.
	For this reason, we only included processing times for cases where a
	chessboard was successfully located in the source image.}
	\label{f:12}
\end{figure}

The only disadvantage of our method is that it is considerably slower than
alternative approaches.  As shown in Figure \ref{f:11}, in some cases our
algorithm is over two times slower than its competitors. The case that was
especially difficult for our algorithm was chessboard number five. However, this
is the only case for which the processing time was longer than 5 seconds, which
was defined as the time limit at the beginning of our study. When comparing the
performances of chess piece detection algorithms, one can see that the method by
Danner and Kafafy is an indisputable leader in terms of speed. It is sometimes
up to 10 times faster than other algorithms when it works correctly (cf. Figure
\ref{f:12}). Again, chessboard five was the most challenging case for our
algorithm for chess piece detection. 

\section{Conclusions}

In this paper, we presented a novel approach for chessboard and chess piece
detection.  To improve its performance, we enhanced various computer vision
methods that are utilized during the detection process, such as the chessboard
lattice point detector or line detector with segment merging. When possible, we
attempted to utilize machine learning methods to improve the accuracy of the
classifiers implemented at various steps of the algorithm.  When designing our
methods, we attempted to make them as robust as possible to achieve accurate
results, regardless of the quality of a source image. We were able to develop a
method that is generic and multipurpose. It allows us to analyze damaged
chessboards with deformed edges, images with poor quality, and  chessboards with
certain parts hidden (e.g., board hidden behind a player's fingers).  The only
disadvantage of our method is that it is slower than alternative approaches.
The full source code for our method can be found at our GitHub repository
{\small\url{https://github.com/maciejczyzewski/neural-chessboard}}. \\
Additionally, we agree with Jialin Ding, who stated that the main difficulty in
research on recognizing chessboards and chess pieces is the lack of
comprehensive labeled datasets of images containing chessboards
\cite{dingchessvision}.  Lack of good datasets is often a problem in other
fields of computer science research \cite{Wasik_2013,Prejzendanc_2016}.  For
this reason, in addition to sharing the source code of our method, we also
decided to share the dataset we prepared \cite{Czyzewski_2018}.

We were able to achieve very satisfactory results when locating chessboards,
largely because of the patient tuning of the methods that are utilized during
our process.  In this manner, we implemented a very robust method that performs
much better than methods utilizing default versions of algorithms available from
computer vision libraries (e.g., the basic algorithms implemented in the method
by Danner and Kafafy \cite{danner2015visual}). Basic computer vision tools,
such as line detectors, often fail in advanced analysis because they cannot
identify a full chessboard structure. Therefore, they are not able to find
chessboard positions accurately.

\begin{figure}[H]\centering
	\includegraphics[width=\columnwidth]{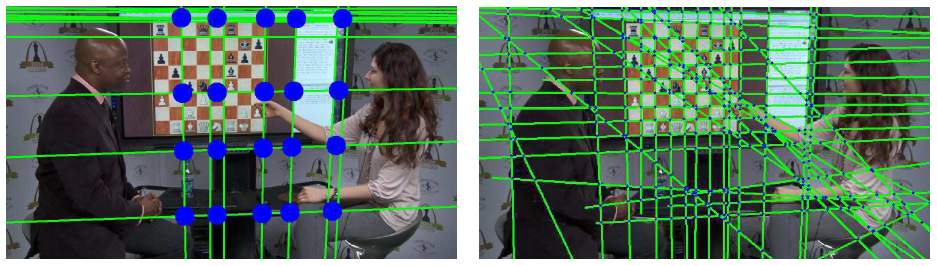}
	\caption{Visual differences between our algorithm (right) and the method by
	Danner and Kafafy \cite{danner2015visual}, which utilizes a single Hough
	transform without segment merging (left). One can see that our method
	generates much more information from an image, which can later be utilized
	to find lattice points and localize a chessboard.}
	\label{f:9}
\end{figure}

For example, our implementation of the straight line detector (SLID algorithm)
attempts to merge many short segments that are detected in an image into longer
segments.  This allows us to find essential lines that could be missed by
algorithms that search only for longer segments. To demonstrate this principle,
we visually compared our method to a method utilizing a single Hough transform
for identifying straight lines in Figure \ref{f:9}.

\begin{figure}[H]\centering
	\includegraphics[width=\columnwidth]{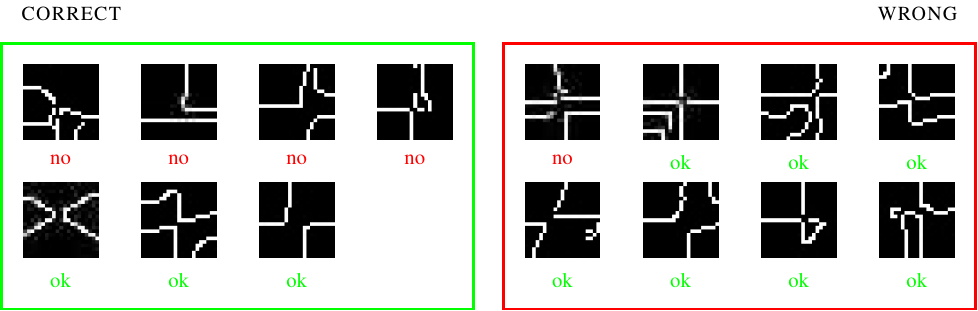}
	\caption{Results of the ChESS detector on randomly selected cases.  The
	cases detected by ChESS as positive are on the left and those detected as
	negative are on the right. Under each chessboard lattice point, there is a
	ground truth label (\textit{no} if it is not a chessboard lattice point and
	\textit{ok} if it is a chessboard lattice point). Here, one can see that the
	ChESS detector only identified $43\%$ of the true positive cases and $13\%$
	of the true negative cases.}
	\label{f:10}
\end{figure}

Another advantage of utilizing computer vision methods boosted by machine
learning can be observed in our lattice points detector. By training the
classifier utilizing a convolutional neural network, we are able to correctly
detect deformed cases of chessboard lattice points, which are poorly recognized
by other algorithms.  For example, in Figure \ref{f:10}, we present the results
of the best-performing alternative method, namely the ChESS classifier. We
observed that this method incorrectly classifies 30\% of deformed points, which
has a negative influence on the accuracy of locating the chessboard because
these points are often key features for analysis.

When our method succeeds in locating a chessboard in an image, it proceeds to
recognize chess pieces. To solve this problem, we utilized a similar approach to
that described by Ding \cite{dingchessvision}. However, we introduced three
major modifications (see Section \ref{ssec:FEN}) that have a substantial impact
on the final results. Boosting our method with the \textit{Stockfish} chess
engine had a significant influence on the accuracy of classification. By
analyzing the probability of different configurations of chess pieces based on
the scores returned by Stockfish, we are able to reject many scenarios that have
a small probability of being real cases. This helps to eliminate cases where
bishops are predicted as pawns, which is a common scenario described by Ding
\cite{dingchessvision}.

In summary, we were able to implement a method that outperforms all alternative
approaches that are currently publicly available. With an accuracy of chess
piece recognition close to $95\%$, it is the first method that can be utilized
in real-world scenarios to digitize chess tournaments or share records of chess
games between friends. When digitizing an entire game, the accuracy can be
improved further because the number of differences between two consecutive
states of a chessboard is significantly limited by the rules of chess. The best
confirmation of our method's usefulness is the significant interest that arose
after the publication of our algorithm on GitHub. We have been contacted by
several people who wish to utilize our method in practical applications. If our
method is put into practice, it will be easy to improve it by adding data
received from users to the training datasets utilized to train our classifiers.

\section*{Acknowledgements}

The authors would like to give special thanks for the sharing of materials and
images from various broadcasts: Saint Louis Chess Club, John Bartholomew, and
Tata Steel Chess Tournament.  The authors received consent from the organizers
of these tournaments to utilize screenshots from broadcasts to train algorithms,
visualize results, and provide supplementary materials. Additionally, SW and AL
were supported by the Polish National Center for Research and Development grant
no. LIDER/004/103/L-5/13/NCBR/2014.

% \nocite{*}

\bibliography{refs}

\begin{thebibliography}{10}

\bibitem{acher2016large}
{\sc Acher, M., and Esnault, F.}
\newblock Large-scale analysis of chess games with chess engines: A preliminary
  report, 2016.

\bibitem{arca2005corner}
{\sc Arca, S., Casiraghi, E., and Lombardi, G.}
\newblock Corner localization in chessboards for camera calibration.
\newblock In {\em Proceedings of International Conference on Multimedia, Image
  Processing and Computer Vision (IADAT-micv2005)\/} (2005).

\bibitem{bency2016weakly}
{\sc Bency, A.~J., Kwon, H., Lee, H., Karthikeyan, S., and Manjunath, B.}
\newblock Weakly supervised localization using deep feature maps.
\newblock In {\em European Conference on Computer Vision\/} (2016), Springer,
  pp.~714--731.

\bibitem{bennett2014chess}
{\sc Bennett, S., and Lasenby, J.}
\newblock Chess--quick and robust detection of chess-board features.
\newblock {\em Computer Vision and Image Understanding 118\/} (2014), 197--210.

\bibitem{braje1998illumination}
{\sc Braje, W.~L., Kersten, D., Tarr, M.~J., and Troje, N.~F.}
\newblock Illumination effects in face recognition.
\newblock {\em Psychobiology 26}, 4 (1998), 371--380.

\bibitem{ZIAUL}
{\sc Choudhury, Z.~H.}
\newblock Biometrics security based on face recognition.
\newblock Master's thesis, India, 2013.

\bibitem{CoolThings2016}
{\sc CoolThings}.
\newblock Square off is a robot chess board that can move pieces on its own,
  November 2016.

\bibitem{Cour2003AutonomousCR}
{\sc Cour, T., Lauranson, R., and Vachette, M.}
\newblock Autonomous chess-playing robot.
\newblock {\em Ecole Polytechnique, July\/} (2002).

\bibitem{Czyzewski_2018}
{\sc Czyzewski, M.~A., Laskowski, A., and Wasik, S.}
\newblock Latchess21: dataset of damaged chessboard lattice points (chessboard
  features) used to train laps detector (grayscale/21x21px), 2018.

\bibitem{danner2015visual}
{\sc Danner, C., and Kafafy, M.}
\newblock Visual chess recognition, 2015.

\bibitem{de2010automatic}
{\sc De~la Escalera, A., and Armingol, J.~M.}
\newblock Automatic chessboard detection for intrinsic and extrinsic camera
  parameter calibration.
\newblock {\em Sensors 10}, 3 (2010), 2027--2044.

\bibitem{dingchessvision}
{\sc Ding, J.}
\newblock Chessvision: Chess board and piece recognition.
\newblock Tech. rep., Stanford University, 2016.

\bibitem{Duda1972}
{\sc Duda, R.~O., and Hart, P.~E.}
\newblock Use of the hough transformation to detect lines and curves in
  pictures.
\newblock {\em Commun. ACM 15}, 1 (Jan. 1972), 11--15.

\bibitem{edwards1994portable}
{\sc Edwards, S.~J.}
\newblock Portable game notation specification and implementation guide.
\newblock {\em Retrieved April 4\/} (1994), 2011.

\bibitem{fernandes2008real}
{\sc Fernandes, L.~A., and Oliveira, M.~M.}
\newblock Real-time line detection through an improved hough transform voting
  scheme.
\newblock {\em Pattern recognition 41}, 1 (2008), 299--314.

\bibitem{Galler_1964}
{\sc Galler, B.~A., and Fisher, M.~J.}
\newblock An improved equivalence algorithm.
\newblock {\em Commun. ACM 7}, 5 (May 1964), 301--303.

\bibitem{gao2017dual}
{\sc Gao, F., Huang, T., Wang, J., Sun, J., Hussain, A., and Yang, E.}
\newblock Dual-branch deep convolution neural network for polarimetric {SAR}
  image classification.
\newblock {\em Applied Sciences 7}, 5 (2017), 447.

\bibitem{hamid2016lsm}
{\sc Hamid, N., and Khan, N.}
\newblock Lsm: perceptually accurate line segment merging.
\newblock {\em Journal of Electronic Imaging 25}, 6 (2016), 061620.

\bibitem{harris1988combined}
{\sc Harris, C., and Stephens, M.}
\newblock A combined corner and edge detector.
\newblock In {\em Alvey vision conference\/} (1988), vol.~15, Citeseer,
  pp.~10--5244.

\bibitem{jassim2013hybridization}
{\sc Jassim, F.~A., and Altaani, F.~H.}
\newblock Hybridization of otsu method and median filter for color image
  segmentation, 2013.

\bibitem{kanchibailchess}
{\sc Kanchibail, R., Suryaprakash, S., and Jagadish, S.}
\newblock Chess board recognition.
\newblock Not published in journal, 2016.

\bibitem{koray2016computer}
{\sc Koray, C., and Sumer, E.}
\newblock A computer vision system for chess game tracking.
\newblock In {\em 21st Computer Vision Winter Workshop, Rimske Toplice,
  Slovenia\/} (2016).

\bibitem{Larson2018}
{\sc Larson, C.}
\newblock China's massive investment in artificial intelligence has an
  insidious downside.
\newblock {\em Science\/} (feb 2018).

\bibitem{Leonard1990}
{\sc Leonard, J., Durrant-Whyte, H., and Cox, I.}
\newblock Dynamic map building for autonomous mobile robot.
\newblock In {\em {IEEE} International Workshop on Intelligent Robots and
  Systems, Towards a New Frontier of Applications\/} (jul 1990), {IEEE}.

\bibitem{Li2004}
{\sc Li, Q., Zheng, N., and Cheng, H.}
\newblock Springrobot: A prototype autonomous vehicle and its algorithms for
  lane detection.
\newblock {\em {IEEE} Transactions on Intelligent Transportation Systems 5}, 4
  (dec 2004), 300--308.

\bibitem{lu2015cannylines}
{\sc Lu, X., Yao, J., Li, K., and Li, L.}
\newblock Cannylines: A parameter-free line segment detector.
\newblock In {\em Image Processing (ICIP), 2015 IEEE International Conference
  on\/} (2015), IEEE, pp.~507--511.

\bibitem{Marciniak2013}
{\sc Marciniak, T., Chmielewska, A., Weychan, R., Parzych, M., and Dabrowski,
  A.}
\newblock Influence of low resolution of images on reliability of face
  detection and recognition.
\newblock {\em Multimedia Tools and Applications 74}, 12 (jul 2013),
  4329--4349.

\bibitem{Matuszek2011GambitAA}
{\sc Matuszek, C., Mayton, B., Aimi, R., Deisenroth, M.~P., Bo, L., Chu, R.,
  Kung, M., LeGrand, L., Smith, J.~R., and Fox, D.}
\newblock Gambit: An autonomous chess-playing robotic system.
\newblock {\em 2011 IEEE International Conference on Robotics and Automation\/}
  (2011), 4291--4297.

\bibitem{Mietchen_2018}
{\sc Mietchen, D., Wodak, S., Wasik, S., Szostak, N., and Dessimoz, C.}
\newblock Submit a topic page to plos computational biology and wikipedia.
\newblock {\em PLOS Computational Biology 14}, 5 (05 2018), 1--4.

\bibitem{Pomerleau1996}
{\sc Pomerleau, D., and Jochem, T.}
\newblock Rapidly adapting machine vision for automated vehicle steering.
\newblock {\em {IEEE} Expert 11}, 2 (apr 1996), 19--27.

\bibitem{Prejzendanc_2016}
{\sc Prejzendanc, T., Wasik, S., and Blazewicz, J.}
\newblock Computer representations of bioinformatics models.
\newblock {\em {Current Bioinformatics} 11}, 5 (2016), 551--560.

\bibitem{reza2004realization}
{\sc Reza, A.~M.}
\newblock Realization of the contrast limited adaptive histogram equalization
  (clahe) for real-time image enhancement.
\newblock {\em Journal of VLSI signal processing systems for signal, image and
  video technology 38}, 1 (2004), 35--44.

\bibitem{sen2010gradient}
{\sc Sen, D., and Pal, S.~K.}
\newblock Gradient histogram: Thresholding in a region of interest for edge
  detection.
\newblock {\em Image and Vision Computing 28}, 4 (2010), 677--695.

\bibitem{Shortis_2015}
{\sc Shortis, M.}
\newblock Calibration techniques for accurate measurements by underwater camera
  systems.
\newblock {\em Sensors 15}, 12 (2015), 30810--30826.

\bibitem{Soh1997}
{\sc Soh, L.}
\newblock Robust recognition of calibration charts.
\newblock In {\em 6th International Conference on Image Processing and its
  Applications\/} (1997), {IEE}.

\bibitem{Stark_2000}
{\sc Stark, J.~A.}
\newblock Adaptive image contrast enhancement using generalizations of
  histogram equalization.
\newblock {\em IEEE Transactions on Image Processing 9}, 5 (May 2000),
  889--896.

\bibitem{Szostak_2016}
{\sc Szostak, N., Wasik, S., and Blazewicz, J.}
\newblock {Hypercycle}.
\newblock {\em {PLOS} Computational Biology 12}, 4 (apr 2016), e1004853.

\bibitem{tam2008automatic}
{\sc Tam, K.~Y., Lay, J.~A., and Levy, D.}
\newblock Automatic grid segmentation of populated chessboard taken at a lower
  angle view.
\newblock In {\em Computing: Techniques and Applications, 2008. DICTA'08.
  Digital Image\/} (2008), IEEE, pp.~294--299.

\bibitem{Tarjan_1975}
{\sc Tarjan, R.~E.}
\newblock Efficiency of a good but not linear set union algorithm.
\newblock {\em J. ACM 22}, 2 (Apr. 1975), 215--225.

\bibitem{tavares1995new}
{\sc Tavares, J. M. R.~S., and Padilha, A. J. M.~N.}
\newblock A new approach for merging edge line segments.
\newblock {\em Proceedings RecPad'95, Aveiro\/} (1995).

\bibitem{Urting2003MarineBlueAL}
{\sc Urting, D., and Berbers, Y.}
\newblock Marineblue: A low-cost chess robot.
\newblock In {\em Robotics and Applications\/} (2003).

\bibitem{Wasik_2015}
{\sc Wasik, S., Fratczak, F., Krzyskow, J., and Wulnikowski, J.}
\newblock {Inferring Mathematical Equations Using Crowdsourcing}.
\newblock {\em {PLOS} {ONE} 10}, 12 (dec 2015), e0145557.

\bibitem{Wasik_2013}
{\sc Wasik, S., Prejzendanc, T., and Blazewicz, J.}
\newblock {ModeLang} - a new approach for experts-friendly viral infections
  modeling.
\newblock {\em {Computational and Mathematical Methods in Medicine} 2013\/}
  (2013), 8.

\bibitem{wiens1996asymptotics}
{\sc Wiens, D.~P.}
\newblock Asymptotics of generalized m-estimation of regression and scale with
  fixed carriers, in an approximately linear model.
\newblock {\em Statistics \& probability letters 30}, 3 (1996), 271--285.

\bibitem{wu2018alive}
{\sc Wu, Q., Zhang, J., Lai, Y.-K., Zheng, J., and Cai, J.}
\newblock Alive caricature from 2d to 3d, 2018.

\bibitem{zhang2000flexible}
{\sc Zhang, Z.}
\newblock A flexible new technique for camera calibration.
\newblock {\em IEEE Transactions on pattern analysis and machine intelligence
  22}, 11 (2000), 1330--1334.

\bibitem{zhao2011automated}
{\sc Zhao, F., Wei, C., Wang, J., and Tang, J.}
\newblock An automated x-corner detection algorithm (axda).
\newblock {\em JSW 6}, 5 (2011), 791--797.

\end{thebibliography}

% \printbibliography

\end{document}